\pdfoutput=1

\documentclass[11pt]{article}
\usepackage{CJKutf8}

\usepackage[hyperref]{acl}
\newcommand*{\affmark}[1][*]{\textsuperscript{#1}}
\usepackage{url}

\usepackage{times}
\usepackage{latexsym}

\usepackage[T1]{fontenc}

\usepackage[utf8]{inputenc}

\usepackage{microtype}

\usepackage{graphicx}
\usepackage{url}
\usepackage{subfigure}
\usepackage{amsmath}
\usepackage[group-separator={,}]{siunitx}
\usepackage{paralist} 
\title{PERT: A New Solution to Pinyin to Character Conversion Task}

\author{
  Jinghui Xiao\affmark[1]  \qquad {\bf Qun Liu}\affmark[1] \qquad {\bf Xin Jiang}\affmark[1] \qquad {\bf Yuanfeng Xiong}\affmark[2] \\ {\bf Haiteng Wu}\affmark[2] \qquad {\bf Zhe Zhang}\affmark[2] \\
  \affmark[1]Huawei Noah's Ark Lab \quad \affmark[2]Huawei Technologies Co., Ltd. \\
  \texttt{\{xiaojinghui4,qun.liu,Jiang.Xin,xiongyuanfeng\}@huawei.com} \\
  \texttt{\{wuhaiteng,zhangzhe88\}@huawei.com} \\
}

\begin{document}
\maketitle
\begin{abstract}
Pinyin to Character conversion (P2C) task is the key task of Input Method Engine (IME) in commercial input software for Asian languages, such as Chinese, Japanese, Thai language and so on. It's usually treated as sequence labelling task and resolved by language model, i.e. n-gram or RNN. However, the low capacity of the n-gram or RNN limits its performance. This paper introduces a new solution named \textbf{PERT} which stands for bidirectional \textbf{P}inyin \textbf{E}ncoder \textbf{R}epresentations from \textbf{T}ransformers. It achieves significant improvement of performance over baselines. Furthermore, we combine PERT with n-gram under a Markov framework, and improve performance further. Lastly, the external lexicon is incorporated into PERT so as to resolve the OOD issue of IME. 
\end{abstract}

\section{Introduction \label{sec:introduction}}

\begin{figure}
	\centering
	\includegraphics[scale=0.65]{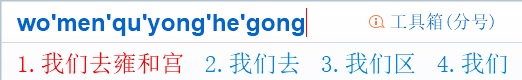}
	\caption{User Types in Chinese via Pinyin in IME}
	\label{fig-ime-input}
\end{figure}

Some Asian languages, such as Chinese, Japanese and Thai language, can not be input directly through standard keyboard. User types in them via commercial input software, such as Microsoft Input Method \citep{DBLP:journals/talip/GaoGLL02}, Sogou Input Method\footnote{\url{https://pinyin.sogou.com/}}, Google Input Method\footnote{\url{https://www.google.com/inputtools/services/features/input-method.html}}, and so on. Pinyin is the official romanization representation for Chinese language. It's natural for user to type in Chinese by pinyin sequence. For example, taking a snapshot from Sogou Input Method as Figure \ref{fig-ime-input} shows, user inputs a pinyin sequence of \textit{wo men qu yong he gong} from the standard keyboard and it is converted into the desired Chinese sentence of ``\begin{CJK}{UTF8}{gbsn}我们去雍和宫\end{CJK} (we are going to Yonghe Lama Temple)''. Therefore,  Pinyin to Character conversion (``P2C'' for short) task is the key task of commercial input software.

\begin{figure}
	\centering
	\includegraphics[scale=0.48]{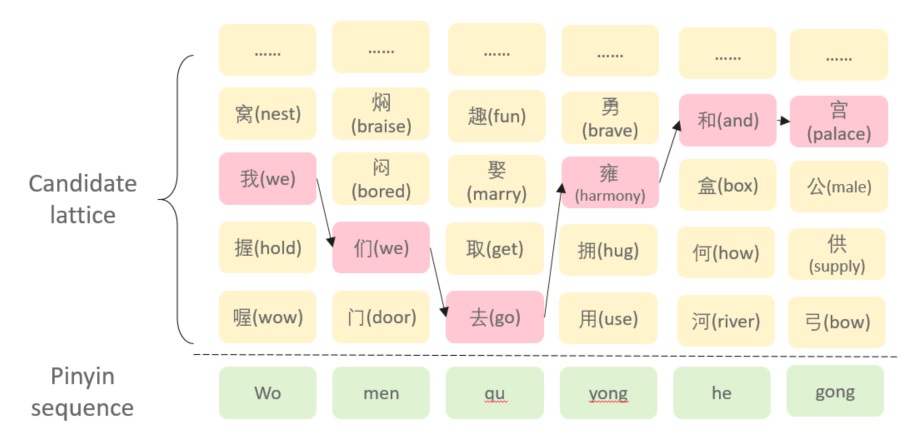}
	\caption{Pinyin to Character Conversion Task}
	\label{fig-ime-viterbi}
\end{figure}

Usually, the P2C task is treated as sequence labelling task and resolved by language model, i.e. n-gram. Specifically, there are three steps as Figure \ref{fig-ime-viterbi} shows. Firstly, the Chinese character candidates are generated according to the input pinyin tokens. All of candidates constitute of a lattice. Secondly, language model provides the probability between characters, i.e. $P(\text{\begin{CJK}{UTF8}{gbsn}们\end{CJK} 
| \begin{CJK}{UTF8}{gbsn}我\end{CJK}})$, for later caculations.  Thirdly, an optimal path with the highest probability in the lattice is found by the Viterbi algorithm. N-gram is the dominant model in the commercial input method software. However, its simplicity and low capacity limit the performance. In the trend of deep learning techniques of recent year, more powerful models, such as RNN \citep{zhaohai2019, DBLP:journals/corr/abs-1810-09309, DBLP:conf/adma/WuHSCL17}, have been applied on the P2C task and achieve substantial improvement. The pre-training language models like BERT-CRF \citep{DBLP:journals/corr/abs-1909-10649, DBLP:conf/bmei/DaiWNLLB19}  are also applied to the sequence labelling task, such as named entity recognition (NER) and outperform n-gram and RNN significantly. 

Inspired by the success of BERT-CRF, we design PERT (bidirectional \textbf{P}inyin \textbf{E}ncoder \textbf{R}epresentations from \textbf{T}ransformers) especially for the P2C task. Instead of the pre-training technique such as BERT-CRF, we train PERT directly on the P2C task on the massive pinyin-character parallel corpus, and achieve substantial improvement of performance over n-gram and RNN. There are three contributions in this paper:

\begin{itemize}
\item We design PERT for the P2C task and achieve substantial improvement of performance. 
\item We combine PERT with n-gram under Markov framework and get further performance improvement. 
\item We incorporate external lexicon in PERT to solve the Out-Of-Domain (``OOD'' for short) issue of IME. 
\end{itemize}

\section{Method \label{sec:method}}

We describe our methodology in this section. Firstly, we present the implement of PERT in Section \ref{sec:pert}. Then we show how to combine PERT with n-gram in Section \ref{sec:modelcombination}. Lastly, we incorporate external lexicon into PERT in Section \ref{sec:pertwithlexicon}. 

\subsection{PERT\label{sec:pert}}

\begin{figure}
	\centering
	\includegraphics[scale=0.52]{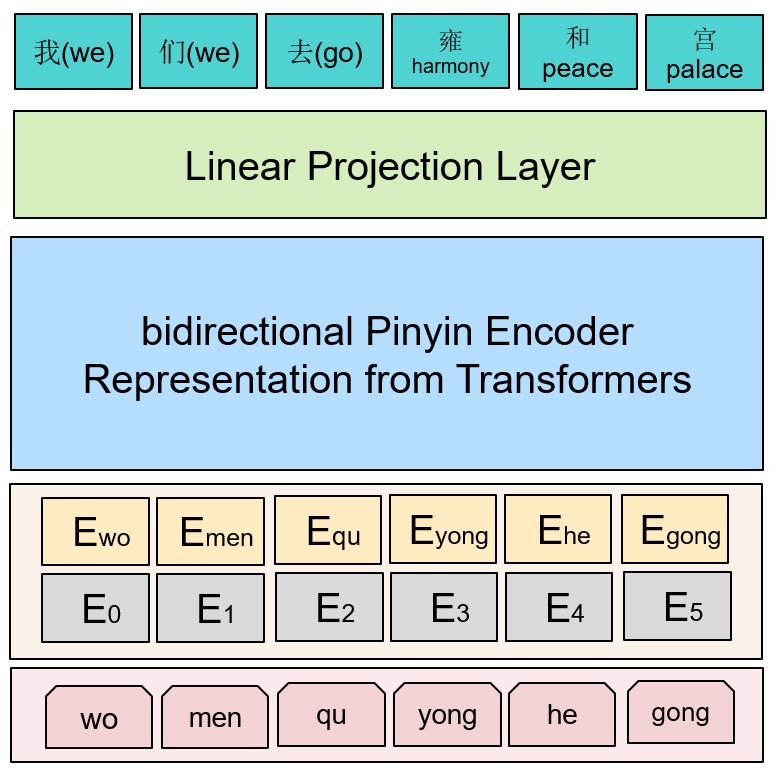}
	\caption{PERT for the P2C Task. \textit{wo} is the input pinyin token. $E_{0}$ is the position embedding and $E_{wo}$ is the token embedding of pinyin. \textit{\begin{CJK}{UTF8}{gbsn}我(we)\end{CJK}} is the predicted Chinese character.}
	\label{fig-pert}
\end{figure}

Figure \ref{fig-pert} shows the network architecture of PERT. There are two differences between BERT and PERT on architecture. Firstly, as shown in the bottom of Figure \ref{fig-pert}, PERT only takes the pinyin token as input since it is designed especially for the P2C task. Whereas, the BERT models for Chinese, such as ChineseBERT \citep{DBLP:conf/acl/SunLSMAHWL20}, take both pinyin and character as input because they are for the common tasks, i.e. Chinese Word Segmentation (CWS), Text Classification (TC) and so on. There is only 410 pinyin tokens (without tone) which is about 2x order of magnitude smaller than Chinese subword (usually about 50k). The scale of embedding layer of PERT is much smaller than BERT. Secondly, there is no segment embedding but only pinyin embedding and position embedding in PERT. PERT is trained directly on the P2C task on the massive pinyin-character parallel corpus, and there is no pre-training process. It doesn't need Next Sentence Prediction (NSP) task, thus no segment embedding.

BERT adopts the pretraining-then-finetune paradigm so as to leverage the massive unlabelled text corpus to supplement limited labelled corpus of target task. However, for the P2C task, we can convert the massive character corpus into the according pinyin corpus accurately and efficiently, and build the massive parallel corpus. Then, PERT can be trained directly on the target P2C task on this corpus. It is called Text-to-Pinyin conversion (``T2P'' for short) task \citep{Text2Pinyin2003} which is the contrary task of the P2C task. The similar tasks are the transliteration task \citep{DBLP:conf/aclnews/KunduPP18} and the phoneme-to-grapheme conversion task \citep{DBLP:journals/corr/abs-1708-01464}. Both of them are resolved with the sequence-to-sequence models. However, the T2P task is much simpler because most of pinyin information can be determined within its lexical context. We can firstly segment the Chinese character sequence into words and then get the correct pinyin token within its word context. The accuracy exceeds $99.9\%$ \citep{Text2Pinyin2003} and there is the open toolkit named PYPinyin \footnote{https://github.com/mozillazg/python-pinyin}.

\begin{figure*}
	\centering
	\includegraphics[scale=0.50]{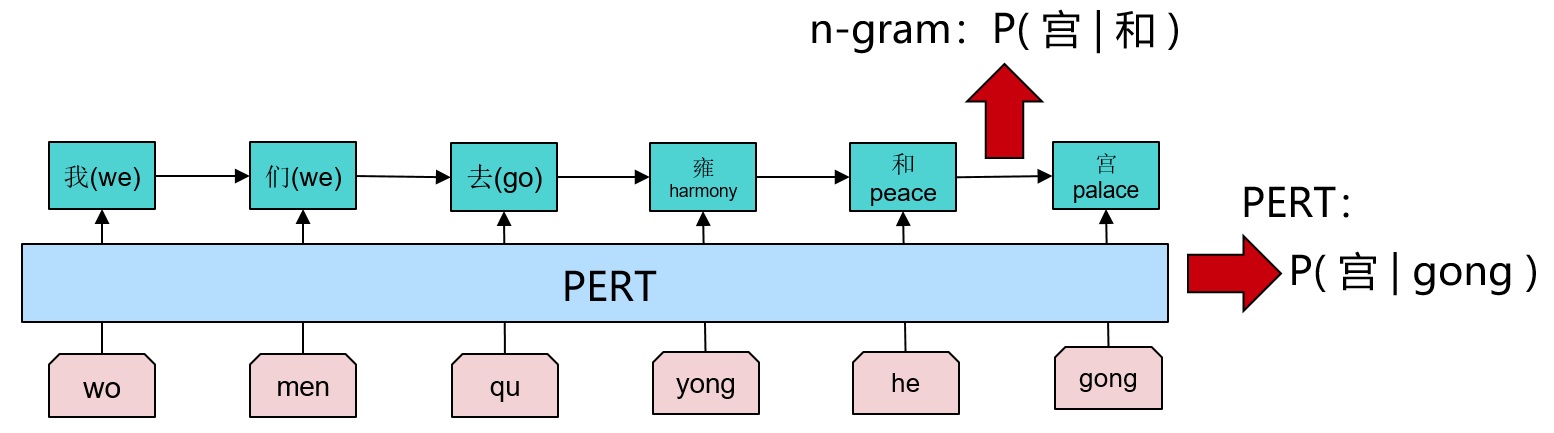}
	\caption{Unify PERT with N-gram Under Markov Framework}
	\label{fig-markov}
\end{figure*}

\begin{figure*}
	\centering
	\includegraphics[scale=0.47]{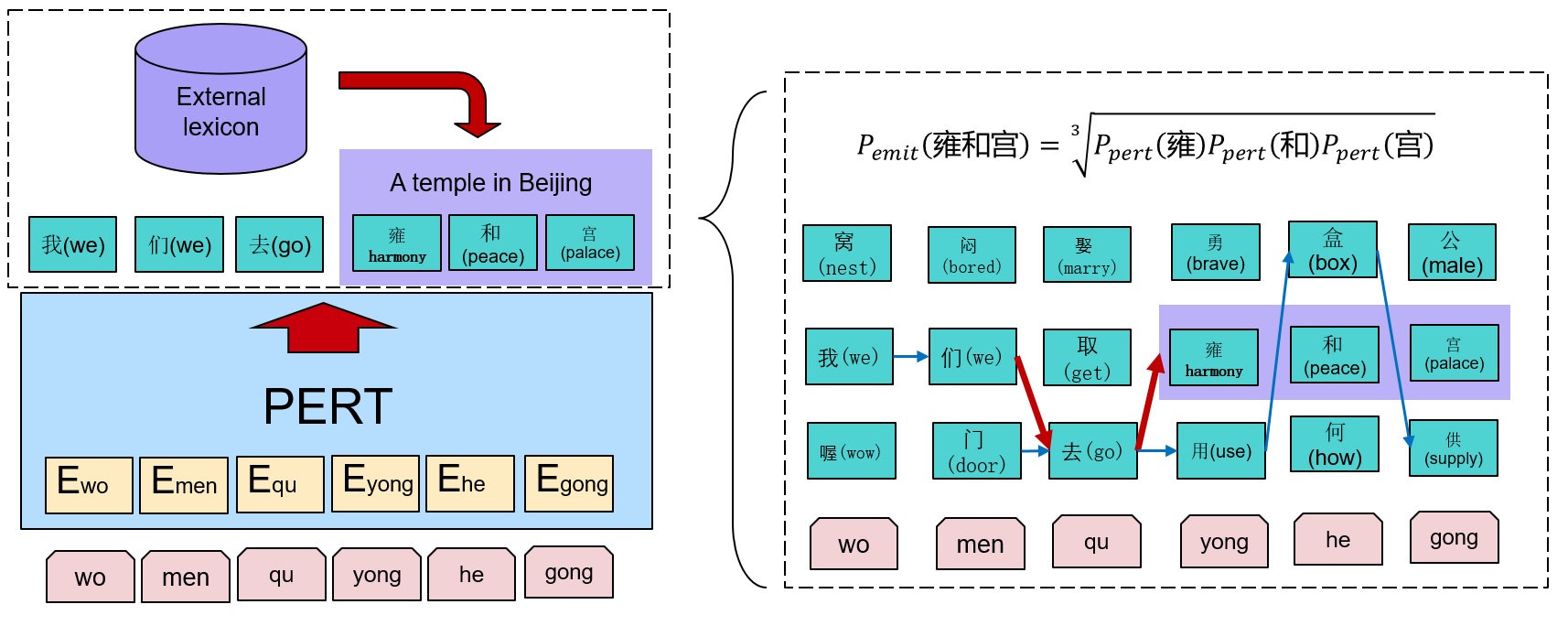}
	\caption{Incorporate Lexicon in PERT}
	\label{fig-addlexicon}
\end{figure*}

\subsection{Combining PERT with N-gram \label{sec:modelcombination}}

In practice, the commercial input software usually installs both n-gram model (for faster installation) and PERT (for higher performance) at the same time.  In this section, we combine these two models together so as to get more capability. We start from the derivation of Bayes rule. Then we unify these two models under Markov's framework. 

In the P2C task, language model estimates the joint conditional probability of $P(c_{1}...c_{n}|y_{1}...y_{n})$, which is the probability of Chinese character sequence $c_{1}...c_{n}$ given the input pinyin sequence $y_{1}...y_{n}$. We can decompose it under the Bayes rule as Formula \ref{formula:bayesian}:


\begin{equation} \label{formula:bayesian}
    \begin{split}
        &P(c_{1}...c_{n}|y_{1}...y_{n})\\
        &=P(c_{1}...c_{n-1},c_{n}|y_{1}...y_{n})\\
        &=P(c_{1}...c_{n-1}|y_{1}...y_{n}) * P(c_{n}|c_{1}...c_{n-1},y_{1}...y_{n})\\
        &\approx P(c_{1}...c_{n-1}|y_{1}...y_{n-1}) * P(c_{n}|c_{1}...c_{n-1},y_{1}...y_{n})\\
        &\approx P(c_{1}...c_{n-1}|y_{1}...y_{n-1})\\
        &\quad \quad * P(c_{n}|c_{1}...c_{n-1}) * P(c_{n}|y_{1}...y_{n})
    \end{split}
\end{equation}

We further simplify these probabilities in two aspects. Firstly, we simplify $P(c_{1}...c_{n-1}|y_{1}...y_{n})$ to $P(c_{1}...c_{n-1}|y_{1}...y_{n-1})$ since $y_{n}$ is \textit{in the future} during user inputs. Secondly, we simplify $P(c_{n}|c_{1}...c_{n-1},y_{1}...y_{n})$ into the product of $P(c_{n}|c_{1}...c_{n-1})$ and $P(c_{n}|y_{1}...y_{n})$ as the last line of Formula \ref{formula:bayesian} shows. As presented in Figure \ref{fig-markov}, $P(c_{n}|c_{1}...c_{n-1})$ can be taken as \textit{the State Transition Probability} of Markov framework, and can be estimated by n-gram.  $P(c_{n}|y_{1}...y_{n})$ is \textit{the Emission Probability} and can be estimated by PERT. $P(c_{1}...c_{n-1}|y_{1}...y_{n-1})$ is \textit{the history} and can be further decomposed in the same way as above. 

Note that in our implementation, n-gram is estimated separately from PERT. Whereas, BERT-CRF estimates BERT and n-gram in an end-to-end way. It's not applicable for BERT-CRF to the P2C task because of two reasons. Firstly, it usually requires different corpus to train n-gram and PERT separately in practice, i.e. the smaller dialog corpus for n-gram whose style is closer to user's real input, but the whole huge corpus (including news, web-text and dialog) for PERT which has more parameters. It can not be trained in an end-to-end way.  Secondly, the target estimation space of BERT-CRF is the whole Chinese characters whose number is more than $6k$. It takes too much GPU memory when calculating the normalization for CRF, which makes it infeasible to train BERT-CRF on a very large corpus\footnote{We adopt Nvidia V100 GPU with 32G memory in our experiments. We encountered the OOM problem when training BERT-CRF at the BERT-base scale. At the BERT-tiny scale, the max batch size was 4, and it took more than five weeks to process about 6 epoch on a training corpus of 300M. On the contrary, the batch size of PERT-tiny was set to 2k and it only took 1.5 hour to finish training 6 epoch on the same corpus. Considering the size of our whole training corpus exceeds 2400M as described later, we thought the cost of training BERT-CRF was unacceptable to us.}.

\subsection{Incorporating External Lexicon \label{sec:pertwithlexicon}}

The commercial IME engine adopts external lexicon to solve the OOD issue on edge device. For example, it adopts the computer lexicon to adapt to the computer domain, adopts the location lexicon to support some location names, i.e. ``\begin{CJK}{UTF8}{gbsn}雍和宫\end{CJK}(Yonghe Lama Temple)'', adopts the user lexicon to fit for user's input custom, and so on. Figure \ref{fig-addlexicon} illustrates our way to incorporate external lexicon in PERT. Specifically there are three steps.  Firstly, we recognize the word item according to the external lexicon, and add it into the lattice of candidates, as shown in the left part of Figure \ref{fig-addlexicon}. Secondly, we estimate the word probability by its component characters according to Formula \ref{formula:yonghegong}. Here we adopt the geometric mean of probability according to the article \citep{DBLP:journals/entropy/Nelson17}. Other metrics can also be tried. Finally, we search an optimal path together with the added word by Viterbi algorithm, as shown by the red arrow of Figure \ref{fig-addlexicon} 's right part. 

\begin{equation} \label{formula:yonghegong}
    \begin{split}
        &P_{emit}(\text{\begin{CJK}{UTF8}{gbsn}雍和宫\end{CJK}})=\\
        &\sqrt[3]{P_{pert}(\text{\begin{CJK}{UTF8}{gbsn}雍\end{CJK}}) * P_{pert}(\text{\begin{CJK}{UTF8}{gbsn}和\end{CJK}}) * P_{pert}(\text{\begin{CJK}{UTF8}{gbsn}宫\end{CJK}})}
    \end{split}
\end{equation}


\section{Experiment}
\label{sec:Experiment}

\subsection{Description of Data Set and Lexicon\label{sec:datadescription}}

As far as we know, there is no benchmark available to the P2C task. So we build our own data set and will make it public to the community later. Table \ref{tab:corpus} describes the detailed information. These articles are collected from some major Chinese news websites, such as Netease, Tencent News and so on. Total 2.6M articles are taken as the training corpus, and another 1k disjoint articles as the test corpus. Besides, two additional corpus from the Baike website and the Society forum \footnote{\url{https://github.com/brightmart/nlp_chinese_corpus}} are chosen to evaluate the OOD performance. All these corpuses are firstly segmented into sentences according to a punctuation list including comma, period, and so on. Secondly, the non-pinyin characters are filtered out, i.e. number, punctuation, English. Thirdly, they are further segmented by a max length ($16$ in our experiment) because user only types a few tokens once a time. Lastly, we convert them into the pinyin sequence by PYPinyin. 

\textit{The Table of General Standard Chinese Characters}\footnote{\url{https://en.wikipedia.org/wiki/Table_of_General_Standard_Chinese_Characters}} containing more than $6k$ characters is taken as the basic lexicon in the experiments. Besides, \textit{Xiandai Hanyu Changyongcibiao (Common words in Contemporary Chinese, ``Common-Words'' for short)} containing $0.1m$ items \citep{modernlexicon:2008} and \textit{Tencent Network Lexicon (`Network-Words'' for short)} published by Tencent AI Lab\footnote{\url{https://ai.tencent.com/ailab/nlp/embedding.html}}  containing $8.8$ million items are taken as the external lexicons.

\begin{table}
\centering
\setlength\tabcolsep{3.5pt}
\begin{tabular}{lccr}
\hline
\textbf{Corpus} & \textbf{\#Articles} & \textbf{\#Chars} & \textbf{\#Disk}\\
\hline
{News-Train} & \num[group-separator={,}]{2603869} & \num[group-separator={,}]{2432585138} & {9.7G}\\
{News-Test} & \num[group-separator={,}]{1000} & \num[group-separator={,}]{926792} & {3.7M}\\
{Baike-Test} & \num[group-separator={,}]{49357} & \num[group-separator={,}]{6238834} & {30M}\\ 
{Society-Test} & \num[group-separator={,}]{68345} & \num[group-separator={,}]{5709450} & {30M}\\ 
\hline
\end{tabular}
\caption{The Detailed Information of Corpus}
\label{tab:corpus}
\end{table}

\subsection{Evaluation Metrics}

In order to evaluate the performance, we use both character-level precision and sentence-level precision. The character-level precision is defined as the ratio that the IME engine converts to the Chinese characters correctly, as described in Formula \ref{formula:precision}.

\begin{small} 
\begin{equation} \label{formula:precision}
Precision_{char}=\frac{\# correct\_converted\_char}{\# total\_converted\_char} 
\end{equation}
\end{small}

Similarly we can define the sentence-level precision which is much stricter since it requires the correctness of whole sequence. Yet it is more meaningful in practice because the input software usually prompts the conversion of whole sequence in its first place and there is shortcut for user to choose that result. 

Besides, we also choose \textit{millisecond per token} to evaluate the latency. 

\begin{table*}
\centering
\small
\setlength{\tabcolsep}{2.5pt}
\begin{tabular}{lcccccc}
\hline
\textbf{Model} & \textbf{\#Parameter} & \textbf{Char\_Precision} & \textbf{Char\_Improvement} & \textbf{Sen\_Precision} & \textbf{Sen\_Improvement} & \textbf{ms/token}\\
\hline
{Bigram} & {7.4M} & {85.10\%} & {NA} & {34.26\%} & {NA} & {1.80}\\
{LSTM} & {4.2M} & {89.71\%} & {4.61\%}$\uparrow$ & {47.87\%} & {13.61\%}$\uparrow$ & {54.63}\\
{PERT-tiny} & {1.3M} & {85.18\%} & {0.08\%}$\uparrow$ & {32.01\%} & {-2.25\%}$\downarrow$ & {0.34}\\
{PERT-mini} & {5.0M} & {90.57\%} & {5.47\%}$\uparrow$ & {48.77\%} & {14.51\%}$\uparrow$ & {0.62}\\
{PERT-small} & {16.5M} & {95.41\%} & {10.31\%}$\uparrow$ & {64.38\%} & {30.12\%}$\uparrow$ & {0.63}\\
{PERT-medium} & {29.1M} & {95.59\%} & {10.49\%}$\uparrow$ & {65.74\%} & {31.48\%}$\uparrow$ & {1.13}\\
{PERT-base} & {91.1M} & {96.59\%} & {11.49\%}$\uparrow$ & {73.31\%} & {39.05\%}$\uparrow$ & {1.71}\\
\hline
\end{tabular}
\caption{\label{tab:py2char}
Performances on the P2C Task. \textit{Char\_Precision} is the character-level precision; \textit{Char\_Improvement} is the improvement of character-level precision; \textit{Sen\_Precision} is the sentence-level precision; \textit{Sen\_Improvement} is the improvement of sentence-level precision. \textit{ms/token} is the millisecond per token. 
}
\end{table*}

\subsection{Baselines and Experiment Settings}

Two kinds of language models, Bigram and LSTM, are taken as baselines. 

\begin{itemize}
\item[$\bullet$] \textbf{Bigram}. Bigram is the de facto model adopted widely in the commercial IME engine. We build it on \textit{The Table of General Standard Chinese Characters} on the training corpus presented in Table \ref{tab:corpus}. No pruning strategy is adopted since the scale of corpus is large enough.

\item[$\bullet$] \textbf{LSTM}. LSTM is reported that gets better performance than Bigram \citep{zhaohai2019,DBLP:journals/corr/abs-1810-09309,DBLP:conf/esann/MalhotraVSA15} in the IME engine. In our implementation, both the embedding size and the hidden size are $256$, and the learning rate is $5e^{-4}$. The batch size is $2k$ and the epoch number is $10$. 
\end{itemize}

For PERT, we follow the specifications of Google from the tiny model to the base model\footnote{\url{https://github.com/google-research/bert}}. We also set its max length to $16$ instead of $512$, so as to be consistent to the training corpus.

\subsection{Experiments on the P2C Task\label{sec:experiment:pinyintocharacterconversion}}

The experimental results are presented in Table \ref{tab:py2char}. Firstly, LSTM outperforms Bigram significantly, which gets $4.61\%$ improvement on character-level precision and $13.61\%$ on sentence-level precision. It confirms the previous conclusion in \citet{DBLP:journals/corr/abs-1810-09309}. Secondly, PERT-tiny gets the comparable performance to Bigram with only about one-sixth parameter number.  And PERT-mini achieves much better performance than Bigram with similar yet smaller number of parameter. It outperforms LSTM too. It proves that PERT has a better network architecture and gets more capability. Lastly, as the scale increases, PERT gets better and better performances. PERT-base achieves $11.49\%$ improvement on character-level precision and $39.05\%$ on sentence-level precision from Bigram. It can be deployed in the resource-rich environment, i.e. on the cloud, so as to provide better services. 

Table \ref{tab:py2char} also presents the inference speed. Bigram is evaluated on Intel(R) Xeon(R) Gold 6561 CPU with 72 cores and 3.00GHz, and others are on Nvidia Tesla V100 GPU with 32M memory. Batch processing is used both for LSTM and PERT. Although LSTM takes much more time than Bigram, it has already been deployed in the real products successfully \citep{zhaohai2019, DBLP:journals/corr/abs-1810-09309}. PERT at all the scales takes much less time than LSTM because it can take fully advantage of parallelism of GPU. Thus, it is reasonable to draw the conclusion that PERT is deployable to real product.

\subsection{Combining PERT with N-gram \label{sec:pinyinbertwithngram}} 

\begin{table}
\centering
\small
\setlength{\tabcolsep}{2.5pt}
\begin{tabular}{lccc}
\hline
\textbf{Model} & \textbf{Char\_Precision} & \textbf{+Bigram} & \textbf{Improved}\\
\hline
{PERT-tiny} & {85.18\%} & {91.28\%} & {6.10\%}$\uparrow$\\
{PERT-mini} & {90.57\%} & {93.68\%} & {3.11\%}$\uparrow$\\
{PERT-small} & {95.41\%} & {96.18\%} & {0.77\%}$\uparrow$\\
{PERT-medium} & {95.59\%} & {96.63\%} & {1.03\%}$\uparrow$\\
{PERT-base} & {96.59\%} & {96.99\%} & {0.40\%}$\uparrow$\\
\hline
\end{tabular}
\caption{\label{tab:plusngram}
Performances on Combining PERT with Bigram.
}
\end{table}

Table \ref{tab:plusngram} presents the experimental results that combines PERT with Bigram as described in Section \ref{sec:modelcombination}.  As we can see, the combined model outperforms the single PERT at every scale of model, which proves that our method can take effective use of the capability of each sub-model and get better performance. Not surprisingly, as the scale increases, PERT becomes more powerful and the gain from the combined model becomes smaller. It's acceptable because the combined model is expected to deploy only on edge device as described in Section \ref{sec:modelcombination}. In the resource-rich environments, we can just deploy PERT as large as possible.

\subsection{Incorporating External Lexicon\label{sec:incorporatingexternallexicon}}

\begin{table*}
\centering
\resizebox{\textwidth}{!}{
\begin{tabular}{lccccc}
\hline
\textbf{Model} & \textbf{Char\_Precision} & \textbf{+Common-Words} & \textbf{Improved} & \textbf{+Network-Words} & \textbf{Improved}\\
\hline
{PERT-tiny} & {83.40\%} & {87.17\%} & {3.77\%}$\uparrow$ & {88.75\%} & {5.35\%}$\uparrow$\\
{PERT-mini} & {88.80\%} & {90.04\%} & {1.24\%}$\uparrow$ & {91.24\%} & {2.44\%}$\uparrow$\\
{PERT-small} & {91.28\%} & {91.86\%} & {0.58\%}$\uparrow$ & {92.75\%} & {1.47\%}$\uparrow$\\
{PERT-medium} & {92.41\%} & {92.88\%} & {0.47\%}$\uparrow$ & {93.57\%} & {1.16\%}$\uparrow$\\
{PERT-base} & {93.68\%} & {94.12\%} & {0.44\%}$\uparrow$ & {94.53\%} & {0.85\%}$\uparrow$\\
\hline
\end{tabular}}
\caption{\label{tab:externallexiconbaike}
Character-level Precision of PERT with External Lexicon on the Baike Domain. \textit{Common-Words} is Xiandai Hanyu Changyongcibiao and \textit{Network-Words} is Tencent Network Lexicon as described in Section \ref{sec:datadescription}.
}
\end{table*}

\begin{table*}
\centering
\resizebox{\textwidth}{!}{
\begin{tabular}{lccccc}
\hline
\textbf{Model} & \textbf{Char\_Precision} & \textbf{+Common-Words} & \textbf{Improved} & \textbf{+Network-Words} & \textbf{Improved}\\
\hline
{PERT-tiny} & {80.26\%} & {84.04\%} & {3.75\%}$\uparrow$ & {86.21\%} & {5.95\%}$\uparrow$\\
{PERT-mini} & {87.13\%} & {87.89\%} & {0.76\%}$\uparrow$ & {89.52\%} & {2.39\%}$\uparrow$\\
{PERT-small} & {90.16\%} & {90.17\%} & {0.01\%}$\uparrow$ & {91.37\%} & {1.21\%}$\uparrow$\\
{PERT-medium} & {91.52\%} & {91.35\%} & {0.17\%}$\downarrow$ & {92.38\%} & {0.86\%}$\uparrow$\\
{PERT-base} & {93.16\%} & {92.97\%} & {0.19\%}$\downarrow$ & {93.61\%} & {0.45\%}$\uparrow$\\
\hline
\end{tabular}}
\caption{\label{tab:externallexiconsociety}
Character-level Precision of PERT with External Lexicon on the Society Domain. \textit{Common-Words} is Xiandai Hanyu Changyongcibiao and \textit{Network-Words} is Tencent Network Lexicon as described in Section \ref{sec:datadescription}.
}
\end{table*}

In this section, we evaluate PERT on two OOD corpus: the Baike corpus and the Society Forum corpus. Then two external lexicons are incorporated into PERT separately so as to improve the performance, as described in Section \ref{sec:pertwithlexicon}. The experimental results are presented in Table \ref{tab:externallexiconbaike} and \ref{tab:externallexiconsociety} respectively. 

Firstly, PERT-tiny gets $83.40\%$ precision on the Baike corpus in Table \ref{tab:externallexiconbaike} and $80.26\%$ on the Society Forum corpus in Table \ref{tab:externallexiconsociety}, whereas it gets $85.18\%$ on the in-domain corpus as shown in Table \ref{tab:py2char}. The performance drops dramatically, which indicates that the OOD issue is very severe to the commercial input software. Secondly, the precision is improved by $3.77\%$ with the Common-words lexicon, and by $5.35\%$ with the Network-words lexicon on the Baike corpus in Table \ref{tab:externallexiconbaike}. Similar results can be found on the Society Forum corpus in Table \ref{tab:externallexiconsociety}. It proves that our method can effectively solve the OOD problem of IME engine. Thirdly, comparing the improvements from two lexicons, the bigger scale of lexicon brings the larger improvement, which also verifies the effectiveness of external lexicon to the OOD problem. Lastly, the gain gradually fade away as the scale of PERT increases. It's acceptable since the external lexicon is only expected to deploy on edge device.

\subsection{Ablation on Scale of Corpus\label{sec:analysisonscalecorpus}}

\begin{figure}
	\centering
	\includegraphics[scale=0.55]{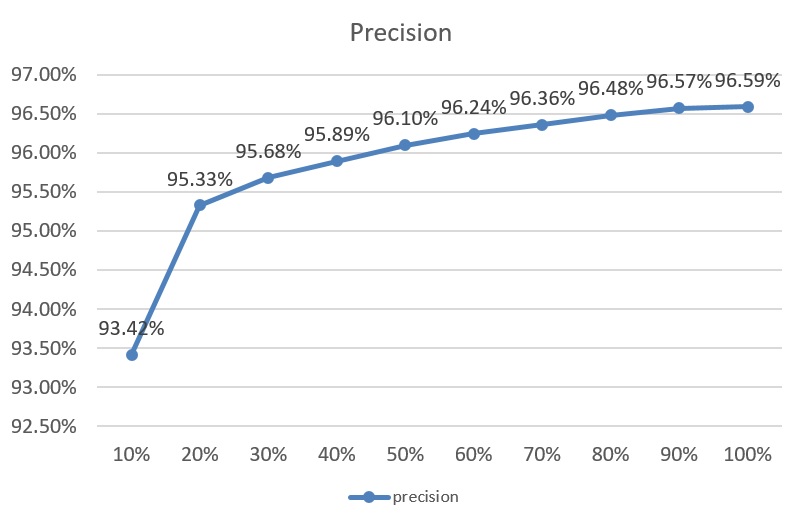}
	\caption{Character-level Precision of PERT on Different Scale of Corpus}
	\label{fig-precision-scale-corpus}
\end{figure}

In the experiments of above sections, we use a large text corpus containing about $2432M$ Chinese characters, whose scale is comparable to Google BERT (about $3300M$ English words). The large scale of language model is always criticized on carbon footprint issue \citep{DBLP:conf/emnlp/Perez-MayosBW21}. In this section, we do the ablation study on the scale of corpus so as to find out whether smaller scale of training corpus is enough to achieve comparable performances. Specifically, we divide the whole corpus into ten pieces, and train PERT-base on them in an accumulated way in the same settings. The results are presented in Figure \ref{fig-precision-scale-corpus}.

The PERT trained on $10\%$ corpus gets only $93.41\%$ precision, which is dramatically lower than  $96.59\%$  of the model trained on the full corpus. Moreover, as the scale increases, the performance increases accordingly. It indicates that it always benefits from the increasing scale of corpus. However, the gain becomes more and more marginal. The PERT on $90\%$ corpus performs almost as good as the full model. 

\section{Related Work}

There are several technical approaches to solve the P2C tasks. It can be treated as sequence labeling task like POS tagging, or as seq2seq task like machine translation, or resolved by pre-trained language model. 

\subsection{Sequence Labeling Task \label{sec:sequencelabelingtask}}

In industry, the P2C task is treated as sequence labeling task. N-gram \citep{DBLP:journals/pami/BahlJM83} is the de facto model adopted widely in commercial input method. Some smoothing methods, such as additive smoothing \citep{additive1825}, interpolation smoothing \citep{interpolation80},  back-off smoothing \citep{katz1987}, Kneser-Ney smoothing \citep{DBLP:conf/icassp/KneserN95}, are adopted to solve the zero-probability problem. Laterly, the exponential models, such as  Maximum Entropy Markov Model (``MEMM'' for short) \cite{DBLP:conf/icml/McCallumFP00} and Conditional Random Field (``CRF'' for short) \citep{DBLP:conf/icml/LaffertyMP01}, are proposed to improve the performance from n-gram.  In recent year, under the trend of deep learning, RNN \citep{DBLP:conf/adma/WuHSCL17} gets more model capacity by capturing longer context.  LSTM \citep{DBLP:journals/corr/abs-1810-09309} is applied on the IME engine and achieves the improvements both in the P2C task and in the candidate prompt task. An incremental selective softmax method is further proposed to speed up the inference. Different from the above works, we adopt the bidirectional encoder representation of Transformer network. Trained on the massive parallel corpus, PERT gets significant performance improvement from n-gram as well as LSTM on the P2C task. 

\subsection{Pre-trained Language Model on P2C Task \label{sec:pretrainedlanguagemodel}}

Recently, the emergence of pre-trained models (PTMs) has brought natural language processing to a new era. BERT \citep{DBLP:conf/naacl/DevlinCLT19} takes the bidirectional encoder representation of transformer networks. It is firstly pre-trained on some self-supervised tasks on the massive unlabelled corpus, such as Masked Language Modeling (``MLM'' for short) and Next Sentence Predicting (``NSP'' for short). Then it is fine-tuned on the target tasks on the small labelled corpus. It achieves the SOTA results on many tasks, including the sequence labeling task. \citet{yuzhanginwheejoe2020} applies pre-trained language model to the P2C task and proposes BERT-P2C. In the experiments, BERT-P2C outperforms other pre-trained models such as ELMO. Different from BERT-P2C, we train PERT directly on the target P2C task by creating the massive labelled corpus instead of the pretrain-then-finetune paradigm, as described in Section \ref{sec:pert}. Besides, comparing with \citet{yuzhanginwheejoe2020} which costs the parameters of two encoders and one decoder of transformers, PERT only takes one encoder whose parameter is much smaller (about one third) than BERT-P2C. Moreover, PERT also outperforms BERT-P2C in the P2C experiments. We describe the experimental results in details in  Appendix \ref{sec:comparisonwithbertp2c}. 

BERT-CRF  \citep{DBLP:journals/corr/abs-1909-10649, DBLP:conf/bmei/DaiWNLLB19} further employs the CRF decoder on the top of BERT encoder, so as to exploit the structure information of target sequence. It is trained in an end-to-end way and achieves the SOTA result on the NER task of Portuguese. Different from BERT-CRF, we train PERT and n-gram separately for two reasons as described in Section \ref{sec:modelcombination}. Firstly, it is required some flexibility when deploying on real product. Secondly, the costs to train BERT-CRF on the whole corpus, especially for the calculation on the normalization of BERT-CRF, are more than that we can afforded. 

ChineseBERT \citep{DBLP:conf/acl/SunLSMAHWL20} exploits some information specific to Chinese, such as glyph and pinyin, so as to enhance its capability on Chinese language. It gets improvement from Google-BERT on a variety of Chinese NLP tasks, such as machine reading comprehension, text classification, named entity recognition and so on. PLOME \citep{DBLP:conf/acl/LiuYYZW20} designs the pre-trained tasks specific to glyph and pinyin in the input text. It is applied on Chinese spelling correction task and gets the SOTA result. For these models, pinyin is only the auxiliary information which has to be input together with glyph and text. It can not be input alone, as the P2C task requires. However, PERT only takes pinyin as input, and it's especially designed for the P2C task.

\subsection{Machine Translation Approach \label{sec:machinetranslation}}

In academic community, the P2C task is also considered as machine translation task \citep{DBLP:conf/iclr/ZhuXWHQZLL20, yuzhanginwheejoe2020,DBLP:conf/acl/ZhangHZ19,zhaohai2019} in which pinyin is taken as source language and Chinese character is target language. It's usually resolved in an encoder-decoder network like Transformer. In practice, the uni-directional decoder is usually the efficiency bottleneck during inference. Then the Non-AutoRegressive (``NAR'' for short) decoder with bi-directional attentions is proposed to solve this problem \citep{DBLP:conf/iclr/Gu0XLS18, DBLP:journals/corr/abs-2004-03227, DBLP:conf/acl/GuK21}. 

In the P2C task, the number of input pinyin tokens is exactly the same as the number of output Chinese characters, which is an explicit constraint that we can utilize. Thus it's more natural to treat it as sequence labeling task as this paper presents. In the future, we can adopt the encoder-decoder solution when we take the continuous letter-level input (without pinyin segmentation before P2C) or the sub-pinyin token input (segmenting input letter sequence into sub-pinyin token sequence) whose lengths are much longer than the length of target character sequence. These models are expected to be more tolerated to input error. We leave them in our future works. Besides, comparing PERT with NAR machine translation system, there is no heavy decoder used in PERT. Thus, the number of PERT parameter is about a half of NAR machine translation system.

\section{Conclusions}

In this paper, we propose PERT for the P2C task which is crucial to the IME engine of commercial input software. In the experiments, PERT outperforms n-gram as well as LSTM significantly. Moreover, we combine PERT with n-gram under Markov framework and get further improvement.  Lastly, we incorporate external lexicon into PERT so as to resolve the OOD issue of IME. 

\section{Future Work}

There are several ideas to try in the future. One interesting idea is to replace the current pinyin-level input with the letter-level input, and convert it into Chinese character sequence by the seq2seq model as discussed in Section \ref{sec:machinetranslation}. The built model is expected to be more tolerated to input error, and makes the IME engine more robust. Online learning is another interesting and important topic. \citet{DBLP:conf/acl/ZhangHZ19} designs an efficient method to update the vocabulary during the user input process, and augment the IME engine with the learned vocabulary effectively. However, \citet{DBLP:conf/acl/ZhangHZ19}  adopts RNN as encoder, which is usually regarded as less capable than Transformer architecture. We are going to incorporate PERT with the online learning method to get further improvement. Lastly, we are going to deploy our model to real product, and compare performance with some commercial input software on real input of user \citep{zhaohai2019,DBLP:conf/acl/HuangLZZ18}.  

\bibliography{anthology,custom}

\begin{thebibliography}{33}
\expandafter\ifx\csname natexlab\endcsname\relax\def\natexlab#1{#1}\fi

\bibitem[{Bahl et~al.(1983)Bahl, Jelinek, and
  Mercer}]{DBLP:journals/pami/BahlJM83}
Lalit~R. Bahl, Frederick Jelinek, and Robert~L. Mercer. 1983.
\newblock \href {https://doi.org/10.1109/TPAMI.1983.4767370} {A maximum
  likelihood approach to continuous speech recognition}.
\newblock \emph{{IEEE} Trans. Pattern Anal. Mach. Intell.}, 5(2):179--190.

\bibitem[{Bu et~al.(2017)Bu, Du, Na, Wu, and
  Zheng}]{DBLP:conf/ococosda/BuDNWZ17}
Hui Bu, Jiayu Du, Xingyu Na, Bengu Wu, and Hao Zheng. 2017.
\newblock \href {https://doi.org/10.1109/ICSDA.2017.8384449} {{AISHELL-1:} an
  open-source mandarin speech corpus and a speech recognition baseline}.
\newblock In \emph{20th Conference of the Oriental Chapter of the International
  Coordinating Committee on Speech Databases and Speech {I/O} Systems and
  Assessment, {O-COCOSDA} 2017, Seoul, South Korea, November 1-3, 2017}, pages
  1--5. {IEEE}.

\bibitem[{Dai et~al.(2019)Dai, Wang, Ni, Li, Li, and
  Bai}]{DBLP:conf/bmei/DaiWNLLB19}
Zhenjin Dai, Xutao Wang, Pin Ni, Yuming Li, Gangmin Li, and Xuming Bai. 2019.
\newblock \href {https://doi.org/10.1109/CISP-BMEI48845.2019.8965823} {Named
  entity recognition using {BERT} bilstm {CRF} for chinese electronic health
  records}.
\newblock In \emph{12th International Congress on Image and Signal Processing,
  BioMedical Engineering and Informatics, {CISP-BMEI} 2019, Suzhou, China,
  October 19-21, 2019}, pages 1--5. {IEEE}.

\bibitem[{Devlin et~al.(2019)Devlin, Chang, Lee, and
  Toutanova}]{DBLP:conf/naacl/DevlinCLT19}
Jacob Devlin, Ming{-}Wei Chang, Kenton Lee, and Kristina Toutanova. 2019.
\newblock \href {https://doi.org/10.18653/v1/n19-1423} {{BERT:} pre-training of
  deep bidirectional transformers for language understanding}.
\newblock In \emph{Proceedings of the 2019 Conference of the North American
  Chapter of the Association for Computational Linguistics: Human Language
  Technologies, {NAACL-HLT} 2019, Minneapolis, MN, USA, June 2-7, 2019, Volume
  1 (Long and Short Papers)}, pages 4171--4186. Association for Computational
  Linguistics.

\bibitem[{Gao et~al.(2002)Gao, Goodman, Li, and
  Lee}]{DBLP:journals/talip/GaoGLL02}
Jianfeng Gao, Joshua Goodman, Mingjing Li, and Kai{-}Fu Lee. 2002.
\newblock \href {https://doi.org/10.1145/595576.595578} {Toward a unified
  approach to statistical language modeling for chinese}.
\newblock \emph{{ACM} Trans. Asian Lang. Inf. Process.}, 1(1):3--33.

\bibitem[{Gu et~al.(2018)Gu, Bradbury, Xiong, Li, and
  Socher}]{DBLP:conf/iclr/Gu0XLS18}
Jiatao Gu, James Bradbury, Caiming Xiong, Victor O.~K. Li, and Richard Socher.
  2018.
\newblock \href {https://openreview.net/forum?id=B1l8BtlCb} {Non-autoregressive
  neural machine translation}.
\newblock In \emph{6th International Conference on Learning Representations,
  {ICLR} 2018, Vancouver, BC, Canada, April 30 - May 3, 2018, Conference Track
  Proceedings}. OpenReview.net.

\bibitem[{Gu and Kong(2021)}]{DBLP:conf/acl/GuK21}
Jiatao Gu and Xiang Kong. 2021.
\newblock \href {https://doi.org/10.18653/v1/2021.findings-acl.11} {Fully
  non-autoregressive neural machine translation: Tricks of the trade}.
\newblock In \emph{Findings of the Association for Computational Linguistics:
  {ACL/IJCNLP} 2021, Online Event, August 1-6, 2021}, volume {ACL/IJCNLP} 2021
  of \emph{Findings of {ACL}}, pages 120--133. Association for Computational
  Linguistics.

\bibitem[{Huang et~al.(2018)Huang, Li, Zhang, and
  Zhao}]{DBLP:conf/acl/HuangLZZ18}
Yafang Huang, Zuchao Li, Zhuosheng Zhang, and Hai Zhao. 2018.
\newblock \href {https://doi.org/10.18653/v1/P18-4024} {Moon {IME:}
  neural-based chinese pinyin aided input method with customizable
  association}.
\newblock In \emph{Proceedings of {ACL} 2018, Melbourne, Australia, July 15-20,
  2018, System Demonstrations}, pages 140--145. Association for Computational
  Linguistics.

\bibitem[{Jelinek and Mercer(1980)}]{interpolation80}
F.~Jelinek and R.~L Mercer. 1980.
\newblock Interpolated estimation of markov source parameters from sparse data.
\newblock In \emph{Proceedings of the Workshop on Pattern Recognition in
  Practice, North-Holland, Amsterdam,}, pages 381--397. The Netherland.

\bibitem[{Kasner et~al.(2020)Kasner, Libovick{\'{y}}, and
  Helcl}]{DBLP:journals/corr/abs-2004-03227}
Zdenek Kasner, Jindrich Libovick{\'{y}}, and Jindrich Helcl. 2020.
\newblock \href {http://arxiv.org/abs/2004.03227} {Improving fluency of
  non-autoregressive machine translation}.
\newblock \emph{CoRR}, abs/2004.03227.

\bibitem[{Katz(1987)}]{katz1987}
S.~M Katz. 1987.
\newblock Esitmation of probabilities from sparse data for the language model
  component of a speech recognizer.
\newblock \emph{IEEE Transactions on Acoustics, Speech and Signal Processin},
  35:400--401.

\bibitem[{Kneser and Ney(1995)}]{DBLP:conf/icassp/KneserN95}
Reinhard Kneser and Hermann Ney. 1995.
\newblock \href {https://doi.org/10.1109/ICASSP.1995.479394} {Improved
  backing-off for m-gram language modeling}.
\newblock In \emph{1995 International Conference on Acoustics, Speech, and
  Signal Processing, {ICASSP} '95, Detroit, Michigan, USA, May 08-12, 1995},
  pages 181--184. {IEEE} Computer Society.

\bibitem[{Kundu et~al.(2018)Kundu, Paul, and Pal}]{DBLP:conf/aclnews/KunduPP18}
Soumyadeep Kundu, Sayantan Paul, and Santanu Pal. 2018.
\newblock \href {https://doi.org/10.18653/v1/w18-2411} {A deep learning based
  approach to transliteration}.
\newblock In \emph{Proceedings of the Seventh Named Entities Workshop, NEWS@ACL
  2018, Melbourne, Australia, July 20, 2018}, pages 79--83. Association for
  Computational Linguistics.

\bibitem[{Lafferty et~al.(2001)Lafferty, McCallum, and
  Pereira}]{DBLP:conf/icml/LaffertyMP01}
John~D. Lafferty, Andrew McCallum, and Fernando C.~N. Pereira. 2001.
\newblock Conditional random fields: Probabilistic models for segmenting and
  labeling sequence data.
\newblock In \emph{Proceedings of the Eighteenth International Conference on
  Machine Learning {(ICML} 2001), Williams College, Williamstown, MA, USA, June
  28 - July 1, 2001}, pages 282--289. Morgan Kaufmann.

\bibitem[{Laplace(1825)}]{additive1825}
P.~S. Laplace. 1825.
\newblock \emph{Philosophical Essay on Probabilities}, 5 edition.
\newblock Springer Verlag.

\bibitem[{Li and Su(2008)}]{modernlexicon:2008}
Xingjian Li and Xinchun Su. 2008.
\newblock \emph{Xiandai Hanyu changyongcibiao}.
\newblock The Commercial Press, Beijing.

\bibitem[{Liu et~al.(2021)Liu, Yang, Yue, Zhang, and
  Wang}]{DBLP:conf/acl/LiuYYZW20}
Shulin Liu, Tao Yang, Tianchi Yue, Feng Zhang, and Di~Wang. 2021.
\newblock \href {https://doi.org/10.18653/v1/2021.acl-long.233} {{PLOME:}
  pre-training with misspelled knowledge for chinese spelling correction}.
\newblock In \emph{Proceedings of the 59th Annual Meeting of the Association
  for Computational Linguistics and the 11th International Joint Conference on
  Natural Language Processing, {ACL/IJCNLP} 2021, (Volume 1: Long Papers),
  Virtual Event, August 1-6, 2021}, pages 2991--3000. Association for
  Computational Linguistics.

\bibitem[{Malhotra et~al.(2015)Malhotra, Vig, Shroff, and
  Agarwal}]{DBLP:conf/esann/MalhotraVSA15}
Pankaj Malhotra, Lovekesh Vig, Gautam Shroff, and Puneet Agarwal. 2015.
\newblock \href
  {http://www.elen.ucl.ac.be/Proceedings/esann/esannpdf/es2015-56.pdf} {Long
  short term memory networks for anomaly detection in time series}.
\newblock In \emph{23rd European Symposium on Artificial Neural Networks,
  {ESANN} 2015, Bruges, Belgium, April 22-24, 2015}.

\bibitem[{McCallum et~al.(2000)McCallum, Freitag, and
  Pereira}]{DBLP:conf/icml/McCallumFP00}
Andrew McCallum, Dayne Freitag, and Fernando C.~N. Pereira. 2000.
\newblock Maximum entropy markov models for information extraction and
  segmentation.
\newblock In \emph{Proceedings of the Seventeenth International Conference on
  Machine Learning {(ICML} 2000), Stanford University, Stanford, CA, USA, June
  29 - July 2, 2000}, pages 591--598. Morgan Kaufmann.

\bibitem[{Meng and Zhao(2019)}]{zhaohai2019}
Zhen Meng and Hai Zhao. 2019.
\newblock \href {http://arxiv.org/abs/1909.01063} {A smart sliding chinese
  pinyin input method editor for touchscreen devices}.

\bibitem[{Nelson(2017)}]{DBLP:journals/entropy/Nelson17}
Kenric~P. Nelson. 2017.
\newblock \href {https://doi.org/10.3390/e19060286} {Assessing probabilistic
  inference by comparing the generalized mean of the model and source
  probabilities}.
\newblock \emph{Entropy}, 19(6):286.

\bibitem[{P{\'{e}}rez{-}Mayos et~al.(2021)P{\'{e}}rez{-}Mayos, Ballesteros, and
  Wanner}]{DBLP:conf/emnlp/Perez-MayosBW21}
Laura P{\'{e}}rez{-}Mayos, Miguel Ballesteros, and Leo Wanner. 2021.
\newblock \href {https://doi.org/10.18653/v1/2021.emnlp-main.118} {How much
  pretraining data do language models need to learn syntax?}
\newblock In \emph{Proceedings of the 2021 Conference on Empirical Methods in
  Natural Language Processing, {EMNLP} 2021, Virtual Event / Punta Cana,
  Dominican Republic, 7-11 November, 2021}, pages 1571--1582. Association for
  Computational Linguistics.

\bibitem[{Peters et~al.(2017)Peters, Dehdari, and van
  Genabith}]{DBLP:journals/corr/abs-1708-01464}
Ben Peters, Jon Dehdari, and Josef van Genabith. 2017.
\newblock \href {http://arxiv.org/abs/1708.01464} {Massively multilingual
  neural grapheme-to-phoneme conversion}.
\newblock \emph{CoRR}, abs/1708.01464.

\bibitem[{Souza et~al.(2019)Souza, Nogueira, and
  de~Alencar~Lotufo}]{DBLP:journals/corr/abs-1909-10649}
F{\'{a}}bio Souza, Rodrigo~Frassetto Nogueira, and Roberto de~Alencar~Lotufo.
  2019.
\newblock \href {http://arxiv.org/abs/1909.10649} {Portuguese named entity
  recognition using {BERT-CRF}}.
\newblock \emph{CoRR}, abs/1909.10649.

\bibitem[{Sun et~al.(2021)Sun, Li, Sun, Meng, Ao, He, Wu, and
  Li}]{DBLP:conf/acl/SunLSMAHWL20}
Zijun Sun, Xiaoya Li, Xiaofei Sun, Yuxian Meng, Xiang Ao, Qing He, Fei Wu, and
  Jiwei Li. 2021.
\newblock \href {https://doi.org/10.18653/v1/2021.acl-long.161} {Chinesebert:
  Chinese pretraining enhanced by glyph and pinyin information}.
\newblock In \emph{Proceedings of the 59th Annual Meeting of the Association
  for Computational Linguistics and the 11th International Joint Conference on
  Natural Language Processing, {ACL/IJCNLP} 2021, (Volume 1: Long Papers),
  Virtual Event, August 1-6, 2021}, pages 2065--2075. Association for
  Computational Linguistics.

\bibitem[{Wang and Zhang(2015)}]{DBLP:journals/corr/WangZ15e}
Dong Wang and Xuewei Zhang. 2015.
\newblock \href {http://arxiv.org/abs/1512.01882} {{THCHS-30} : {A} free
  chinese speech corpus}.
\newblock \emph{CoRR}, abs/1512.01882.

\bibitem[{Wu et~al.(2017)Wu, Haynes, Smith, Chen, and
  Li}]{DBLP:conf/adma/WuHSCL17}
Lin Wu, Michele Haynes, Andrew Smith, Tong Chen, and Xue Li. 2017.
\newblock \href {https://doi.org/10.1007/978-3-319-69179-4\_16} {Generating
  life course trajectory sequences with recurrent neural networks and
  application to early detection of social disadvantage}.
\newblock In \emph{Advanced Data Mining and Applications - 13th International
  Conference, {ADMA} 2017, Singapore, November 5-6, 2017, Proceedings}, volume
  10604 of \emph{Lecture Notes in Computer Science}, pages 225--242. Springer.

\bibitem[{Xu et~al.(2020)Xu, Zhang, and
  Dong}]{DBLP:journals/corr/abs-2003-01355}
Liang Xu, Xuanwei Zhang, and Qianqian Dong. 2020.
\newblock \href {http://arxiv.org/abs/2003.01355} {Cluecorpus2020: {A}
  large-scale chinese corpus for pre-training language model}.
\newblock \emph{CoRR}, abs/2003.01355.

\bibitem[{Yao et~al.(2018)Yao, Shu, Li, Ohtsuki, and
  Nakayama}]{DBLP:journals/corr/abs-1810-09309}
Jiali Yao, Raphael Shu, Xinjian Li, Katsutoshi Ohtsuki, and Hideki Nakayama.
  2018.
\newblock \href {http://arxiv.org/abs/1810.09309} {Real-time neural-based input
  method}.
\newblock \emph{CoRR}, abs/1810.09309.

\bibitem[{Zhang and Laprie(2003)}]{Text2Pinyin2003}
Sen Zhang and Yves Laprie. 2003.
\newblock \href {https://hal.inria.fr/inria-00107717/document} {Text-to-pinyin
  conversion based on contextual knowledge and d-tree for mandarin}.
\newblock In \emph{IEEE International Conference on Natural Language Processing
  and Knowledge Engineering, {NLP-KE} 2003, Beijing, China, 2003}.

\bibitem[{Zhang and Joe(2020)}]{yuzhanginwheejoe2020}
Yu~Zhang and Inwhee Joe. 2020.
\newblock A pre-trained language model for chinese pinyin-to-character task
  based on bert.
\newblock \emph{Thesis for the Master of Science}, pages 197--198.

\bibitem[{Zhang et~al.(2019)Zhang, Huang, and Zhao}]{DBLP:conf/acl/ZhangHZ19}
Zhuosheng Zhang, Yafang Huang, and Hai Zhao. 2019.
\newblock \href {https://doi.org/10.18653/v1/p19-1154} {Open vocabulary
  learning for neural chinese pinyin {IME}}.
\newblock In \emph{Proceedings of the 57th Conference of the Association for
  Computational Linguistics, {ACL} 2019, Florence, Italy, July 28- August 2,
  2019, Volume 1: Long Papers}, pages 1584--1594. Association for Computational
  Linguistics.

\bibitem[{Zhu et~al.(2020)Zhu, Xia, Wu, He, Qin, Zhou, Li, and
  Liu}]{DBLP:conf/iclr/ZhuXWHQZLL20}
Jinhua Zhu, Yingce Xia, Lijun Wu, Di~He, Tao Qin, Wengang Zhou, Houqiang Li,
  and Tie{-}Yan Liu. 2020.
\newblock \href {https://openreview.net/forum?id=Hyl7ygStwB} {Incorporating
  {BERT} into neural machine translation}.
\newblock In \emph{8th International Conference on Learning Representations,
  {ICLR} 2020, Addis Ababa, Ethiopia, April 26-30, 2020}. OpenReview.net.

\end{thebibliography}
\bibliographystyle{acl_natbib}

\begin{table*}
\centering
\begin{tabular}{ll}
\hline
\textbf{Case 1} & \\
{Pinyin Sequence} & {zhi shi \textbf{dai} le yi fu mo jing} \\
{Bigram Result} &
{\begin{CJK}{UTF8}{gbsn}只是\end{CJK}\begin{CJK}{UTF8}{gkai}带\end{CJK}\begin{CJK}{UTF8}{gbsn}了一副墨镜\end{CJK} (Only \textbf{bring} a sunglasses)} \\
{LSTM Result} & {\begin{CJK}{UTF8}{gbsn}只是\end{CJK}\begin{CJK}{UTF8}{gkai}戴\end{CJK}\begin{CJK}{UTF8}{gbsn}了一副墨镜\end{CJK} (Only \textbf{wear} a sunglasses)} \\
\hline
\textbf{Case 2} & \\
{Pinyin Sequence} & {\textbf{shi} wan jia neng gou qing song ying dui} \\
{Bigram Result} & {\begin{CJK}{UTF8}{gkai}是\end{CJK}\begin{CJK}{UTF8}{gbsn}玩家能够轻松应对\end{CJK} (\textbf{Is} player handle it easily)} \\
{LSTM Result} & {\begin{CJK}{UTF8}{gkai}使\end{CJK}\begin{CJK}{UTF8}{gbsn}玩家能够轻松应对\end{CJK} (\textbf{Enable} player handle it easily)}\\
\hline
\textbf{Case 3} & \\
{Pinyin Sequence} & {\textbf{ji} qing liang \textbf{you} shi shang} \\
{Bigram Result} & {\begin{CJK}{UTF8}{gkai}及\end{CJK}\begin{CJK}{UTF8}{gbsn}清凉\end{CJK}\begin{CJK}{UTF8}{gkai}有\end{CJK}\begin{CJK}{UTF8}{gbsn}很时尚\end{CJK} (\textbf{And} cool \textbf{has} fashion)} \\
{LSTM Result} & {\begin{CJK}{UTF8}{gkai}既\end{CJK}\begin{CJK}{UTF8}{gbsn}清凉\end{CJK}\begin{CJK}{UTF8}{gkai}又\end{CJK}\begin{CJK}{UTF8}{gbsn}很时尚\end{CJK} (\textbf{Both} cool \textbf{and} fashion)} \\
\hline
\end{tabular}
\caption{\label{tab:casebigramandrnn}
Case Studies on Comparing Bigram with LSTM
}
\end{table*}

\begin{table*}
\centering
\begin{tabular}{ll}
\hline
\textbf{Case 1} & \\
{Pinyin Sequence} & {\textbf{er shi} gong gong wei sheng fang kong de wen ti} \\
{LSTM Result} & {\begin{CJK}{UTF8}{gkai}二十\end{CJK}\begin{CJK}{UTF8}{gbsn}公共卫生防控的问题\end{CJK}} \\ {} & {(\textbf{Twenty problems} of public health prevention and control)} \\
{PERT Result} & {\begin{CJK}{UTF8}{gkai}二是\end{CJK}\begin{CJK}{UTF8}{gbsn}公共卫生防控的问题\end{CJK}} \\ {} & {(\textbf{The second problem} is about public health prevention and control)}\\
\hline
\textbf{Case 2} & \\
{Pinyin Sequence} & {zhe ge shi pin hen \textbf{chong qing}} \\
{LSTM Result} & {\begin{CJK}{UTF8}{gbsn}这个视频很\end{CJK}\begin{CJK}{UTF8}{gkai}宠情\end{CJK} (This video is very \textbf{pampering})} \\
{PERT Result} & {\begin{CJK}{UTF8}{gbsn}这个视频很\end{CJK}\begin{CJK}{UTF8}{gkai}重庆\end{CJK} (This video is very \textbf{chongqing-style})}\\
\hline
\textbf{Case 3} & \\
{Pinyin Sequence} & {man wei zhong \textbf{gu yi fa shi} ban yan zhe tai li hai le} \\
{LSTM Result} & {\begin{CJK}{UTF8}{gbsn}漫威中\end{CJK}\begin{CJK}{UTF8}{gkai}故意发誓\end{CJK}\begin{CJK}{UTF8}{gbsn}扮演者太厉害了\end{CJK}} \\ {} & {(The role of \textbf{intentional pledge} in Marvel is awesome)} \\
{PERT Result} & {\begin{CJK}{UTF8}{gbsn}漫威中\end{CJK}\begin{CJK}{UTF8}{gkai}古一法师\end{CJK}\begin{CJK}{UTF8}{gbsn}扮演者太厉害了\end{CJK}} \\ {} & {(The role of \textbf{Ancient One} in Marvel is awesome)}\\
\hline
\end{tabular}
\caption{\label{tab:casernnandpinyinbert}
Case Studies on Comparing LSTM with PERT
}
\end{table*}

\appendix
\section{Comparison with BERT-P2C \label{sec:comparisonwithbertp2c}}

We compare the performance of PERT with BERT-P2C \citep{yuzhanginwheejoe2020} in this section. Like BERT, BERT-P2C is firstly pre-trained on the CLUECorpus2020 corpus \citep{DBLP:journals/corr/abs-2003-01355} which contains about 100G texts. Then the pinyin-character parallel corpus is extracted from two open-source Mandarin speech corpus, named AISHELL-1\citep{DBLP:conf/ococosda/BuDNWZ17} and  THCHS-30\citep{DBLP:journals/corr/WangZ15e}. They are combined together and further divided into the training corpus and the test corpus. BERT-P2C is then fine-tuned on the P2C task on the training corpus. Evaluated on the test corpus, BERT-P2C achieves $94.6\%$ character-level precision.

We choose PERT-base for comparison. We continue the settings in Section \ref{sec:Experiment} and train PERT directly on the training corpus containing about 9.7G News texts. Evaluated directly on the test corpus of BERT-P2C, it gets $93.06\%$ character-level precision which is slightly lower than $94.6\%$ of BERT-P2C. Then, we continue to train PERT on the training corpus of BERT-P2C. The batch size is $32$, the learning rate is $3e^{-4}$ and the epoch number is $20$. At this time, PERT achieves $97.77\%$ precision which is much higher than BERT-P2C. Besides, considering the scale of training corpus of PERT is much less (about one tenth) than that of BERT-P2C, and the parameter number of PERT is also much less (about one third) than BERT-P2C, PERT can be trained more efficiently and effectively. In summary, it is proven that PERT is the better choice to the P2C task than BERT-P2C is. 

\section{Error Analysis \label{sec:casestudy}}

In this section, we choose some cases from the experiments and get in-depth analysis. 

Table \ref{tab:casebigramandrnn} compares Bigram with LSTM. In Case 1, the conversion of \textit{dai} is determined by the last word of ``\begin{CJK}{UTF8}{gbsn}墨镜\end{CJK} (sunglasses)''. Bigram can not capture such a long context and it gets an incorrect conversion of ``\begin{CJK}{UTF8}{gbsn}带\end{CJK} (bring)''. Whereas, LSTM can model the whole sentence and get the correct conversion of ``\begin{CJK}{UTF8}{gbsn}戴\end{CJK} (wear)''. It's similar for the pinyin \textit{shi} in Case 2 which can be predicted by the word of ``\begin{CJK}{UTF8}{gbsn}应对\end{CJK} (handle)''. Moreover, LSTM can also recognize some fix collocation like ``\begin{CJK}{UTF8}{gbsn}既...又...\end{CJK} (both...and...)'', and get the result correctly, as shown in Case 3. In a word, Bigram can only captures local context, which limits its performance on the P2C task. However, LSTM can take the whole sequence into consideration, and get improvement from Bigram. 

Table \ref{tab:casernnandpinyinbert} compares LSTM with PERT. In Case 1, both conversions are grammatically correct. However, from the semantic point of view, the result of PERT makes more sense. In Case 2, the city name of ``\begin{CJK}{UTF8}{gbsn}重庆(chongqing)\end{CJK}'' acts as an adjective which means \textit{the style or fashion like Chongqing} in the context. PERT can capture these semantic information and get the correct conversion, whereas LSTM can not. Moreover, we also find that PERT can handle named entity in sentence much better than LSTM, as shown in Case 3. All these cases prove that PERT can better understand the semantic information in sentence and thus get better performance.

\end{document}